\documentclass[10pt,twocolumn,letterpaper]{article}

\usepackage[pagenumbers]{cvpr} %

\usepackage{epsfig}
\usepackage{graphicx}
\usepackage{amsmath}
\usepackage{amssymb}
\usepackage{booktabs}

\usepackage{times}
\usepackage{float}
\usepackage{algorithm}
\usepackage[noend]{algpseudocode}

\usepackage{color}
\usepackage{multirow}
\usepackage{diagbox}
\usepackage{mathrsfs}
\usepackage{xcolor} 

\usepackage{stfloats}
\usepackage{dsfont}
\usepackage{nicefrac}
\usepackage{enumitem}
\usepackage{colortbl}
\usepackage{marvosym}
\usepackage[export]{adjustbox}
\usepackage{pifont}
\usepackage{threeparttable}

\usepackage{listings}
\newcommand{\gr}{\rowcolor[rgb]{0.95,0.97,1.0}}
\newcommand{\rt}{\textcolor[rgb]{0.75,0.25,0.25}}
\newcommand{\gt}{\textcolor[rgb]{0.25,0.75,0.25}}

\usepackage[pagebackref=true,breaklinks,colorlinks,citecolor=cyan,linkcolor=cyan,urlcolor=cyan]{hyperref}

\usepackage[capitalize]{cleveref}
\crefname{section}{Sec.}{Secs.}
\Crefname{section}{Section}{Sections}
\Crefname{table}{Table}{Tables}
\crefname{table}{Tab.}{Tabs.}

\def\cvprPaperID{*****} %
\def\confName{CVPR}
\def\confYear{2023}

\begin{document}

\title{Rethinking Performance Gains in Image Dehazing Networks}

\author {
	Yuda Song{\footnotemark[2]}
	\quad Yang Zhou{\footnotemark[2]}
	\quad Hui Qian
	\quad Xin Du{\textsuperscript \Letter}\\
	Zhejiang University, Hangzhou, China \\
	{\tt\small \{syd,yang\_zhou,qianhui,duxin\}@zju.edu.cn}
}

\maketitle

\renewcommand{\thefootnote}{\fnsymbol{footnote}}
\footnotetext[2]{Equal contribution.}
\renewcommand{\thefootnote}{\arabic{footnote}}

\begin{abstract}
    Image dehazing is an active topic in low-level vision, and many image dehazing networks have been proposed with the rapid development of deep learning.
    Although these networks' pipelines work fine, the key mechanism to improving image dehazing performance remains unclear.
    For this reason, we do not target to propose a dehazing network with fancy modules; rather, we make minimal modifications to popular U-Net to obtain a compact dehazing network.
    Specifically, we swap out the convolutional blocks in U-Net for residual blocks with the gating mechanism, fuse the feature maps of main paths and skip connections using the selective kernel, and call the resulting U-Net variant gUNet.
    As a result, with a significantly reduced overhead, gUNet is superior to state-of-the-art methods on multiple image dehazing datasets.
    Finally, we verify these key designs to the performance gain of image dehazing networks through extensive ablation studies.
    We share all of our code and data
    at \href{https://github.com/IDKiro/gUNet}{https://github.com/IDKiro/gUNet}.
    \vspace{-0.5em}
\end{abstract}

\section{Introduction}

Image dehazing has long been a topic of discussion in low-level vision, as haze is a common atmospheric phenomenon that compromises public safety and personal life.
To get better-looking photos in bad weather is the main use case for image dehazing for personal use, and these features are frequently included in photo editing software.
Besides, image dehazing is generally viewed to be a prerequisite before high-level vision tasks for industrial use.
For example, it is feasible to use the image dehazing module as a pre-processing module to increase the robustness of autonomous driving systems.
All this makes image dehazing one of the representative tasks of low-level vision and receives increasing attention from researchers.

The goal of image dehazing aims to restore a haze-free image from a hazy image.
Researchers often use atmospheric scattering models~\cite{mccartney1976optics,nayar1999vision,narasimhan2002vision} to characterize the degradation process for hazy images:
\begin{equation}
    \label{eq:haze}
    I = J(x) t(x) + A (1 - t(x)),
\end{equation}
where $I$ is the hazy image, $J$ is the latent haze-free image, $A$ is the global atmospheric light, and $t$ is the medium transmission map.
The transmission map $t$ can further formulated as $t(x) = e^{-\beta d(x)}$, where $\beta$ is the scattering coefficient of the atmosphere, and $d$ is the scene depth.

\begin{figure}[t]
    \centering
    \includegraphics[width=1.0\columnwidth]{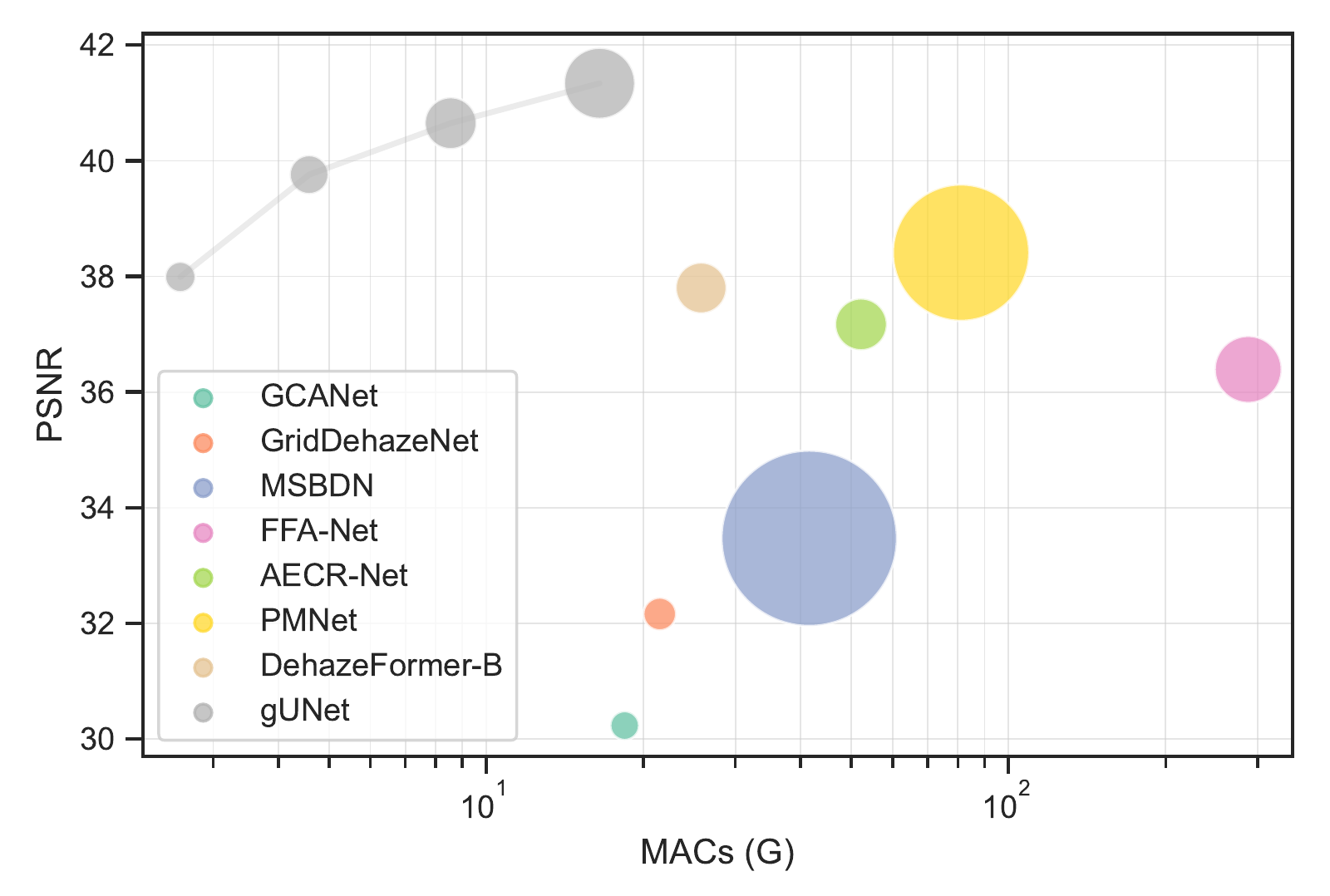}
    \caption{
        Comparison of gUNet with other image dehazing methods on the SOTS indoor set.
        The size of the dots indicates the \#Param, and MACs are shown with the logarithmic axis.
    }
    \label{fig:intro}
\end{figure}

Generally, image dehazing methods can be roughly categorized into prior-based methods and learning-based methods.
Early image dehazing methods are mainly based on handcraft priors ~\cite{he2010single,zhu2015fast,fattal2014dehazing,berman2016non}. 
Recently, many researchers are looking into image dehazing networks~\cite{cai2016dehazenet,li2017aod,qin2020ffa,song2022vision} because of the superior performance of the learning-based methods.
Although many impressive image dehazing networks have been proposed, the keys to improving the dehazing performance are still unclear. 
In this paper, we focus more on exploring the key design of image dehazing networks.
We review well-known image dehazing networks and found two moments of significant performance gain.
The first is GCANet~\cite{chen2019gated} and GridDehazeNet~\cite{liu2019griddehazenet}, which both introduce to extract multi-scale information and no longer predict $A$ and $t(x)$ separately but to predict the latent haze-free image or the residual between the haze-free image and the hazy image.
The second is FFA-Net~\cite{qin2020ffa}, which proposed a deep network that introduces the pixel attention module and channel attention module.
Other works also make sound modification suggestions.
However, these modifications only give modest performance gains but make the network more complicated and hard to deploy.

We try to create a minimal implementation that incorporates these key designs based on the observations mentioned above.
Specifically, we first use the classical U-Net~\cite{long2015fully} with local residuals~\cite{he2016deep} and global residual~\cite{zhang2017beyond} as our base architecture to extract multi-scale information.
Then, we use depth-wise separable convolutional layers~\cite{howard2017mobilenets,sandler2018mobilenetv2} to aggregate spatial information and transform features efficiently.
Further, we assign extraction of global information to the module based on the SK module~\cite{li2019selective} that dynamically fuses feature maps from different paths channel-wise.
Finally, we introduce the gating units in the convolutional blocks, and these gating units act as the pixel attention module and nonlinear activation function.
As a result, two key modules are proposed, that is, a residual block with the gating mechanism called gConv block and a fusion module with the channel attention mechanism called SK Fusion layer.
Here we name our model gUNet because it is a simple U-Net variant with gating.

We evaluate the performance of gUNet on four image dehazing datasets, and for each dataset, we train four variants with different depths.
The experimental results show that gUNet can substantially outperform contemporaneous methods with significantly lower overhead.
Fig.~\ref{fig:intro} shows the comparison of gUNet with other image dehazing methods on the most frequently used SOTS indoor set.
It can be seen that the four variants of gUNet are in the chart's upper left corner, meaning that they outperform all image dehazing methods with lower computational cost.
Specifically, the tiny model gUNet-T outperforms DehazeFormer-B, relying on 10\% computational cost and 32\% parameters, while the small model gUNet-S outperforms PMNet, using only 5.6\% computational cost and 7.4\% parameters.
More importantly, we performed extensive ablation studies on all four datasets to verify these key designs. 
The experimental results show that our proposed module can achieve consistent performance gains on image dehazing.

\section{Related Work}

Image dehazing methods are roughly divided into two categories: prior-based methods and learning-based methods.
Early image dehazing methods are generally based on handcraft priors and produce images with good visibility.
With the development of deep learning, learning-based methods have dominated image dehazing in recent years.

\noindent\textbf{Prior-based image dehazing.}
The prior-based methods often start from statistical differences between the haze-free and hazy images.
Dark channel prior (DCP)~\cite{he2010single} may be one of the most famous prior-based methods.
DCP argues that for outdoor images, at least one color channel in the local patches of the haze-free image has an intensity close to 0. 
Based on this proposal, DCP estimates the transmission map from the hazy image, together with a simple global atmospheric light estimation, to obtain a vivid dehazed image.
Besides, color-lines~\cite{fattal2014dehazing} and haze-lines~\cite{berman2016non} analyze the linear relationship of channels of the haze-free in RGB color space from different viewpoints and thus estimate the transmission map.
In contrast, color attenuation prior (CAP)~\cite{zhu2015fast} analyzes the relationship between the S-V channel differences and scene depth in HSV color space for hazy images and thus derives the transmission map.
Recently, rank-one prior~\cite{liu2021rank} considers the transmission map close to a rank-1 matrix, which can be derived by computing the image's projection on the unified radiance.
However, because these priors are based on empirical statistics, they may lead to unrealistic results when the scenes do not satisfy these priors.
The most well-known case is that DCP will perform poorly in large white regions (\emph{e.g.}, sky).

\noindent\textbf{Learning-based image dehazing.}
Learning-based methods are often generalized from large-scale datasets to estimate haze-free images.
DehazeNet~\cite{cai2016dehazenet} and MSCNN~\cite{ren2016single} are the pioneers of learning-based methods for image dehazing.
Because their architecture is very shallow and the global atmospheric light is still estimated by the conventional method, they do not show performance far exceeding prior-based methods.
DCPDN~\cite{zhang2018densely} instead uses two sub-networks to estimate the transmission map and global atmospheric light, respectively.
By rewriting Eq.(\ref{eq:haze}), AOD-Net~\cite{li2017aod} only predicts a single output and gets good dehazing performance with only an extremely lightweight backbone network.
GridDehazeNet~\cite{liu2019griddehazenet} proposes a grid-like network architecture with a large number of skip connections. 
More importantly, it proposes learning to restore the image directly since it can obtain better performance.
After this, image dehazing networks tend to estimate the haze-free image or the residual between the haze-free image and the hazy image.
For the moment, most learning-based methods tend to design complicated network architectures to improve the expressive capability of the network, some introduce attention mechanisms~\cite{dong2020physics,deng2020hardgan,qin2020ffa,wu2021contrastive}, while some introduce multi-scale features~\cite{chen2019gated,dong2020multi,wang2021eaa}.
Very recently, a few ViT-based~\cite{dosovitskiy2020image} dehazing networks are proposed, such as HyLoG-ViT~\cite{zhao2021hybrid}, DehazeFormer~\cite{song2022vision}, and DeHamer~\cite{guo2022dehamer}.
However, the pipeline of ViT is far more complex than that of CNN, leading to a considerable hindrance in its deployment.
Going a different route from the previous methods, we do not aim at a complicated network architecture but propose a U-Net variant with minimal changes to facilitate the deployments. 
And our proposed method can outperform previous methods, even though it seems quite simple.

\section{gUNet}

\begin{figure*}[!t]
    \centering
    \includegraphics[width=1.0\textwidth]{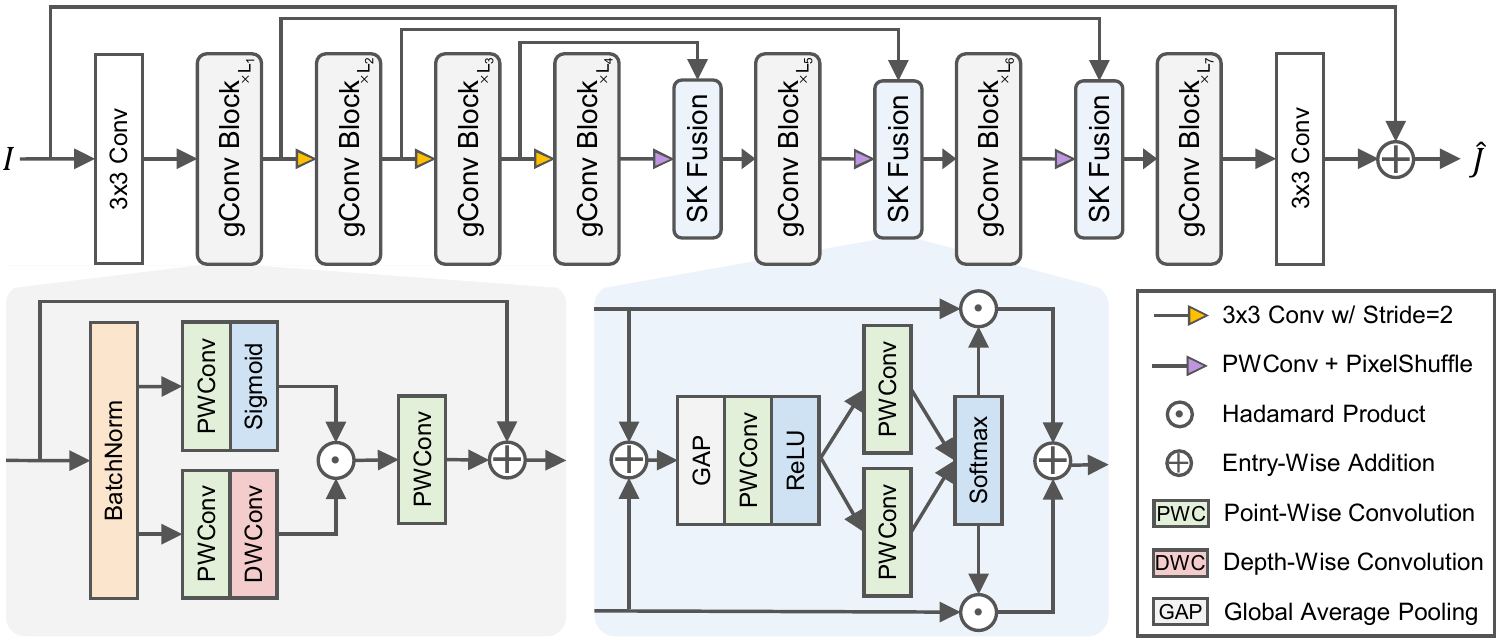}
    \caption{
        Our proposed gUNet is a simple U-Net variant.
        Compared to the conventional U-Net architecture, gUNet uses gConv blocks and SK fusion layers to replace the original convolutional blocks and concatenation fusion layers.
    }
    \label{fig:overview}
\end{figure*}

Figure.~\ref{fig:overview} shows the overall architecture of the gUNet.
Our model, gUNet, can be seen as a 7-stage U-Net variant, whose each stage consists of a stack of proposed gConv blocks.
Besides, gUNet does not adopt the strategy of using concatenation following the convolutional layer to fuse the skip connection and the main path, which U-Net adopts.
Instead, we propose using the SK Fusion module to fuse the feature maps from different paths dynamically.
Given the image pair $\{ I(x), J(x) \}$, we let gUNet predict the global residual~\cite{zhang2017beyond,chen2019gated,qin2020ffa}, \emph{i.e.}, $R(x)=\hat{J}(x) - I(x)$, then compute the $L_1$ loss to train our models.

\subsection{Motivation}

We first describe the motivation of gUNet, which is based on key designs from previous works.
The first is the extraction of multi-scale information, and we use the classical U-Net~\cite{long2015fully} as our base architecture, which produces feature maps of different sizes and thus extracts multi-scale features.
Then, we add local residuals~\cite{he2016deep} to the convolutional blocks and global residual~\cite{zhang2017beyond} to the network.
To make the network deeper without significantly increasing the number of parameters and computational cost, we use the depth-wise separable convolution~\cite{howard2017mobilenets,sandler2018mobilenetv2} to aggregate information and transform features efficiently.
Now the key to network design is how to use attention mechanisms to improve the network's expressive capability.
We recall Eq.(\ref{eq:haze}) and find that atmospheric light $A$ is a shared global variable, while $t(x)$ is a location-dependent local variable.
In FFA-Net, the channel attention module is the only module that can effectively extract global information.
We believe that the channel attention mechanism efficiently extracts the information necessary to encode $A$, which is one of the reasons why FFA-Net works.
However, although the computational cost of the channel attention module is minor, the number of parameters and latency it introduces is non-negligible.
We consider that estimating $A$ should be a simple task since there are many methods~\cite{he2010single,cai2016dehazenet,ren2016single,zhang2018densely} allocate most of the computational resources to estimate $t(x)$ but use lightweight modules to estimate $A$.
Thus we only assign this task to the fusion module based on SK module~\cite{li2019selective} that dynamic fuses feature maps from different paths.
Correspondingly, the pixel attention module aims to make the network pay more attention to informative features.
We find that the gate mechanism in GLU~\cite{dauphin2017language,shazeer2020glu} plays a similar role.
To this end, we introduce the gating mechanism in the convolution block and make it act as the pixel attention module and nonlinear activation function.

\subsection{gConv Block}
Our gConv is mainly based on gMLP~\cite{liu2021pay} and GLU~\cite{dauphin2017language,shazeer2020glu}.
Let $x$ be the feature map, we first normalize it using BatchNorm~\cite{ioffe2015batch} via $\hat{x} = \operatorname{BatchNorm}(x)$.
For inference, BatchNorm uses the exponential moving averages of statistics tracked on the train set. 
It can be merged with the adjacent linear layers, more in line with the requirements of lightweight networks.
Moreover, BatchNorm does not have the drawback of LayerNorm~\cite{ba2016layer} that breaks the spatial correlation mentioned in DehazeFormer~\cite{song2022vision}.
\begin{equation}
    \begin{split}
        x_1 &= \operatorname{Sigmoid}(\operatorname{PWConv_1}(\hat{x})), \\
        x_2 &= \operatorname{DWConv}(\operatorname{PWConv_2}(\hat{x})), \\
    \end{split}
\end{equation}
where PWConv means point-wise convolutional layer and DWConv means depth-wise convolutional layer.
Then we use $x_1$ as a gating signal for $x_2$, which is then projected using another PWConv, and the output is summed with the identity shortcut $x$, which can be formulated as:
\begin{equation}
    y = x + \operatorname{PWConv_3}(x_1 \cdot x_2).
\end{equation}

The use of gating mechanisms to improve the expressive capability of networks is not a new idea in other image restoration tasks~\cite{zamir2021restormer,tu2022maxim,chen2022simple}.
The most similar work to ours is NAFNet~\cite{chen2022simple}, considering that both of us do not use conventional nonlinear activation functions such as ReLU and GELU but rely only on the gating mechanism to achieve nonlinearity.
In contrast, NAFNet uses a bilinear variant of GLU (\emph{i.e.}, without any nonlinear activation function), and we use the original version of GLU (\emph{i.e.}, using sigmoid as the gating function).

\subsection{SK Fusion}

The SK fusion layer is a simple modification of the SK module~\cite{li2019selective}.
Similar ideas can be found in MIRNet~\cite{zamir2020mirnet,zamir2022learning} and DehazeFormer.
Let the two feature maps be $x_1$ and $x_2$, where $x_1$ is the feature maps from the skip connection, and $x_2$ is the feature maps from the main path.
We first use a PWConv layer $f(\cdot)$ to project $x_1$ to $\hat{x}_1=f(x_1)$, which is not illustrated in Figure~\ref{fig:overview}.
We use the global average pooling $\operatorname{GAP}(\cdot)$, MLP (PWConv-ReLU-PWConv) $\mathcal{F}_{mlp} (\cdot)$, softmax function and split operation to obtain the fusion weights:
\begin{equation}
    \{a_1, a_2\}=\operatorname{Split}(\operatorname{Softmax}(\mathcal{F}_{mlp}(\operatorname{GAP}\left(\hat{x}_1 + x_2 \right))).
\end{equation}
Finally, we fuse $\hat{x}_1, x_2$ via $y = a_1 \hat{x}_1 + a_2 x_2$.
To reduce the number of parameters, the two PWConv layers of MLP are the dimensionality-reduction and dimensionality-increasing layers, which are consistent with the conventional channel attention mechanism~\cite{hu2018squeeze}.

\subsection{Training Strategy}

We used several techniques that are not often employed in training image restoration networks for training gUNet. 

\noindent\textbf{SyncBN.}
In standard implementations of BatchNorm, normalization is performed separately on each GPU, which results in a smaller normalization batch size~\cite{wu2021rethinking} than the mini-batch size in data-parallel training.
Therefore, we use SyncBN~\cite{peng2018megdet}, an implementation of BatchNorm that enables the normalization batch size up to the mini-batch size.
We find that the best performance can be achieved at the normalization batch size of 16 and larger, so we only enable SyncBN when the normalization batch size is smaller than 16.
SyncBN introduces additional overhead because it shares the mean and variance between multiple GPUs.

\noindent\textbf{FrozenBN.}
FrozenBN is a constant affine transform since it uses previously computed population statistics at training time.
Once the tracked population statistics are fixed, then the optimization benefits from the normalization layer are lost, but it can keep the train-test consistent.
We enable FrozenBN in the last few epochs because the learning rate is low, and there is no need to worry about gradient explosion.
As a result, we find that FrozenBN can significantly improve the model performance when the total number of training epochs or the normalization batch size is small.

\noindent\textbf{Mixed Precision Training.}
Mixed precision training enables low precision training on some layers during training to reduce the computational cost and memory usage without degrading the model's performance.
Even on some high-level vision tasks, mixed precision training can slightly improve the accuracy of the model~\cite{he2019bag}.
We enable mixed precision training to reduce the training time and increase the mini-batch size.

\noindent\textbf{Warmup.}
The linear warmup is often used in high-level vision tasks to help optimize.
Because our initial learning rate is relatively large and mixed precision training is enabled, we find that the model may produce NaN during training.
To lower the risk of collapse, we apply warmup strategy.

\subsection{Implementation Details}

For simplicity, we set the number of gConv blocks per stage to $\{M,M,M,2M,M,M,M\}$ and the number of channels to $\{N,2N,4N,8N,4N,2N,N\}$, where $M$ is base block number, and $N$ is base channel number.
To verify the scalability of gUNet, we propose four gUNet variants (-T, -S, -B, and -D for tiny, small, basic, and deep, respectively).
We set the width and kernel size $k$ of DWConv of all variants to be the same, specifically, $N = 24$ and $k=5$.
The four variants differ only in-depth, and we set their base block numbers $M$ to $\{2, 4, 8, 16\}$.

We use 4-card RTX-3090 to train our models.
When training, images are randomly cropped to $256 \times 256$ patches.
Considering different datasets have different sample numbers, we set the number of samples per epoch to 16,384 and the total number of epochs to 1,000, whose first 50 epochs are for warmup and the last 200 epochs are for FrozenBN.
In this way, we can exclude the effect of the training iterations and better analyze the differences in ablation studies on different datasets.
Limited by the GPU memory, we set the mini-batch size to $\{128, 128, 64, 32\}$ for \{-T, -S, -B, -D\}, respectively.
For gUNet-D, its normalization batch size is smaller than 16, so we enable SyncBN.
Based on the linear scaling rule~\cite{goyal2017accurate}, we set the initial learning rate to $\{16,16,8,4\}  \times 10^{-4}$ for \{-T, -S, -B, -D\}.
We use AdamW optimizer~\cite{loshchilov2017decoupled} ($\beta_1=0.9,\beta_2=0.999$) with the cosine annealing strategy~\cite{loshchilov2016sgdr} to train the models, where the learning rate gradually decreases from the initial learning rate to $\{16,16,8,4\}  \times 10^{-6}$.

\section{Experiments}

\subsection{Experimental Setup}

We perform our experiments on the RESIDE~\cite{li2018benchmarking}, Haze4K~\cite{liu2021synthetic}, and RS-Haze~\cite{song2022vision} datasets.
For the RESIDE dataset, we follow the setup of FFA-Net~\cite{qin2020ffa}, which contains two main experimental setups, RESIDE-IN and RESIDE-OUT.
Specifically, we use the ITS (13,990 image pairs) and OTS (313,950 image pairs) to train the models and test them on the indoor set (500 image pairs) and the outdoor set (500 image pairs) of the SOTS, respectively.
For the Haze4K dataset, we follow the setup of PMNet~\cite{ye2021perceiving}.
Haze4K contains 4,000 image pairs, of which 3,000 are used for training and the remaining 1,000 for testing. 
Compared to RESIDE, Haze4K mixes images of indoor and outdoor scenes, and the synthesis pipeline is more realistic.
For the RS-Haze dataset, we follow the setup of DehazeFormer~\cite{song2022vision}.
It consists of 54,000 image pairs, of which 51,300 are used for training and the remaining 2,700 for testing. 
RS-Haze is a remote sensing dehazing image dataset with a more monotonous scene, but its haze is highly non-homogeneous, the exact opposite of RESIDE and Haze4K.
All models are trained using their original strategies, and we replicate the best results reported in the previous works.

\subsection{Quantitative Comparison}

\begin{table*}[t]
    \centering
    \caption{
        Quantitative comparison of image dehazing methods trained on different datasets.
        For the compared methods, we use \textbf{bold}, and \underline{underline} to mark the best and second best methods.
        For gUNet, we use \textbf{bold} or \underline{underline} to indicate that the model is better than the best or second best of compared methods.
    }
    \vspace{-0.5em}
    \label{tab:quantitative}
    \begin{center}
        \renewcommand\arraystretch{1.25}
        \resizebox{1.0\linewidth}{!}{
            \begin{threeparttable}
            \begin{tabular}{lccccccccccc}
                \hline
                \multirow{2}*{Methods}            & \multicolumn{2}{c}{RESIDE-IN} & \multicolumn{2}{c}{RESIDE-OUT} & \multicolumn{2}{c}{Haze-4K} & \multicolumn{2}{c}{RS-Haze} & \multicolumn{3}{c}{Overhead} \\
                \cline{2-12}
                                                      & PSNR  & SSIM   & PSNR  & SSIM   & PSNR  & SSIM   & PSNR  & SSIM   & \#Param (M) & \hspace{-0.75em} MACs (G) \hspace{-0.75em} & Latency (ms) \\
                \hline\hline
                DCP~\cite{he2010single}               & 16.62 & 0.818  & 19.13 & 0.815  & 14.01 & 0.760  & 17.86 & 0.734  & -           & -        & -            \\
                \gr DehazeNet~\cite{cai2016dehazenet} & 19.82 & 0.821  & 24.75 & 0.927  & 19.12 & 0.840  & 23.16 & 0.816  & 0.009       & 0.581    & 0.919        \\
                MSCNN~\cite{ren2016single}            & 19.84 & 0.833  & 22.06 & 0.908  & 14.01 & 0.510  & 22.80 & 0.823  & 0.008       & 0.525    & 0.619        \\
                \gr AOD-Net~\cite{li2017aod}          & 20.51 & 0.816  & 24.14 & 0.920  & 17.15 & 0.830  & 24.90 & 0.830  & 0.002       & 0.115    & 0.390        \\
                \hline
                GFN~\cite{ren2018gated}               & 22.30 & 0.880  & 21.55 & 0.844  & -     & -      & 29.24 & 0.910  & \textbf{0.499}       & \textbf{14.94}    & \underline{3.849}        \\
                \gr GCANet~\cite{chen2019gated}       & 30.23 & 0.980  & -     & -      & -     & -      & 34.41 & 0.949  & \underline{0.702}       & \underline{18.41}    & \textbf{3.695}        \\
                GDNet~\cite{liu2019griddehazenet}     & 32.16 & 0.984  & 30.86 & 0.982  & 23.29 & 0.930  & 36.40 & 0.960  & 0.956       & 21.49    & 9.905        \\
                \gr MSBDN~\cite{dong2020multi}        & 33.67 & 0.985  & 33.48 & 0.982  & 22.99 & 0.850  & 38.57 & 0.965  & 31.35       & 41.54    & 13.25        \\
                PFDN~\cite{dong2020physics}           & 32.68 & 0.976  & -     & -      & -     & -      & 36.04 & 0.955  & 11.27       & 50.46    & 4.809        \\
                \gr FFA-Net~\cite{qin2020ffa}         & 36.39 & 0.989  & 33.57 & 0.984  & \underline{26.96} & \underline{0.950}  & \underline{39.39} & \underline{0.969}  & 4.456       & 287.8    & 55.91        \\
                AECR-Net~\cite{wu2021contrastive}     & 37.17 & 0.990  & -     & -      & -     & -      & 35.69 & 0.959  & 2.611       & 52.20    & 6.095        \\
                \gr PMNet~\cite{ye2021perceiving}     & \textbf{38.41} & \underline{0.990}  & \underline{34.74} & \textbf{0.985}  & \textbf{33.49} & \textbf{0.980}  & -     & -      & 18.90       & 81.13    & 28.08        \\
                DehazeF-B~\cite{song2022vision}       & \underline{37.84} & \textbf{0.994}  & \textbf{34.95} & \underline{0.984}  & -     & -      & \textbf{39.87} & \textbf{0.971}  & 2.514       & 25.79    & 27.16        \\
                \hline
                \gr gUNet-T                           & \underline{37.99} & \underline{0.993} & {34.52} & {0.983} & \underline{31.60} & \textbf{0.984} & {38.80} & {0.967} & {0.805} & \textbf{2.595} & \textbf{3.391} \\
                \gr gUNet-S                           & \textbf{39.76} & \textbf{0.995} & \textbf{35.17} & \underline{0.984} & \underline{32.18} & \textbf{0.985} & {39.22} & \underline{0.969} & {1.408} & \textbf{4.579} & {5.377} \\
                \gr gUNet-B                           & \textbf{40.65} & \textbf{0.996} & \textbf{36.29} & \textbf{0.986} & \underline{33.12} & \textbf{0.988} & {39.35} & \underline{0.970} & {2.614} & \textbf{8.548} & {9.712} \\
                \gr gUNet-D                           & \textbf{41.34} & \textbf{0.996} & \textbf{36.64} & \textbf{0.986} &    \textbf{33.52} & \textbf{0.988} & \underline{39.70} & \textbf{0.971} & {5.025} & \underline{16.48} & {19.65} \\
                \hline
            \end{tabular}
            \begin{tablenotes}
                \small
                \item[1] MACs and Latency are measured on $256 \times 256$ images using a single RTX 3090.
                \item[2] For a clearer comparison of models' overheads, earlier methods will be excluded.
              \end{tablenotes}
            \end{threeparttable}
        }
    \end{center}
\end{table*}

We quantitatively compare the performance of gUNet and baselines, and the results are shown in TABLE~\ref{tab:quantitative}.
Our proposed gUNet is a better image dehazing network considering both effectiveness and efficiency.
Specifically, our proposed small model (gUNet-S) outperforms all previous methods on the most widely used RESIDE dataset.
And on the Haze-4K and RS-Haze datasets, gUNet also achieves a comparable performance to the state-of-the-art models with higher efficiency.
Finally, all variants of gUnet work well, and we believe it is a scalable method.
For RESIDE dataset, gUNet-S outperforms PMNet on the most widely used indoor set and surpasses DehazeFormer-B on the outdoor set.
For Haze-4K, the performance of the trained model varied widely because of the small sample size of Haze-4K. 
We believe that much of the performance gap between gUNet-D and PMNet stems from the randomness of the experiments.
For RS-Haze, although gUNet-D is more efficient than DehazeFormer-B, DehazeFormer is still the best method.
We believe this is because the images in the RS-Haze dataset contain many repeat scenes that are more suitable for self-attention~\cite{mei2020pyramid}.
Finally, it can be found that gUNet is not fast enough, although it has a compact architecture with low computational costs, which is due to that DWConv's memory access cost is high~\cite{ding2022scaling}.
Fortunately, running deep networks in edge devices is more limited by computing capacity than memory bandwidth.

\subsection{Qualitative Comparison}

\begin{figure*}[t]
    \vspace{-0.4em}
    \centering
    \includegraphics[width=1.0\textwidth]{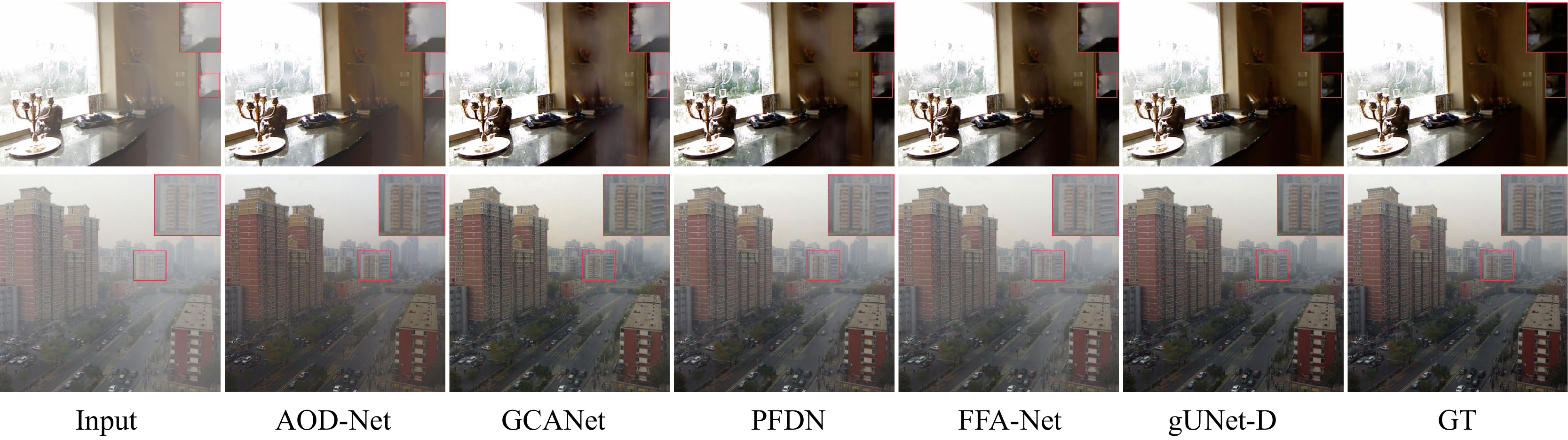}
    \caption{
        Qualitative comparison of image dehazing methods on SOTS, where the first two rows are indoor images, and the last two are outdoor images.
        The first column is the hazy images, and the last column is the corresponding ground truth.
    }
    \label{fig:compare1}
\end{figure*}

We select two representative samples of RESIDE dataset to analyze the performance of each method qualitatively.
Fig.~\ref{fig:compare1} illustrate qualitative comparisons of our gUNet-D with some dehazing networks. 
For the indoor scene, we can find that GCANet, PFDN, and FFA-Net all produce artifacts, while AOD-Net does not produce artifacts, but it has too strong dehazing effort for the wall and too little for the ground.
In contrast, gUNet-D produces no artifacts, and the brightness and tones of each area of its restored image are very close to the ground truth.
Besides, gUNet can better distinguish light-colored objects from dark-colored objects masked by haze, producing more satisfying results accordingly.
For the outdoor scene, gUNet's dehazing results are much closer to a clean image than all the compared methods.
However, the clean images of outdoor scene samples in the synthetic dataset (RESIDE and Haze4K) contain haze, and all trained models cannot remove them, which indicates that the synthetic haze is not realistic enough.

\subsection{Ablation Study}

\begin{table*}[t]
    \renewcommand\arraystretch{1.25}
    \centering
    \caption{
        Ablation study of network architectures and training strategies on different datasets.
        The results marked in \gt{green} or \rt{red} means there is an improvement or degradation compared to the baseline (gUNet-T).
        We train baseline five times and take the best and worst.
    }
    \vspace{-0.5em}
    \label{tab:ablation}
    \begin{center}
        \resizebox{1.0\linewidth}{!}{
            \begin{threeparttable}
            \begin{tabular}{lccccccccccc}
                \hline
                \multirow{2}*{Methods}            & \multicolumn{2}{c}{RESIDE-IN} & \multicolumn{2}{c}{RESIDE-OUT} & \multicolumn{2}{c}{Haze-4K} & \multicolumn{2}{c}{RS-Haze} & \multicolumn{3}{c}{Overhead}\\
                \cline{2-12}
                                                       & PSNR  & SSIM  & PSNR  & SSIM  & PSNR  & SSIM  & PSNR  & SSIM  & \#Param (M) & \hspace{-0.75em} MACs (G) \hspace{-0.75em} & Latency (ms) \\
                \hline\hline
                Baseline - Best                        & 38.11 & 0.993 & 34.61 & 0.983 & 31.68 & 0.984 & 38.93 & 0.968 & 0.805       & 2.595    & 3.391        \\
                \hspace{3.4em} - Worst                 & 37.66 & 0.993 & 34.47 & 0.983 & 31.47 & 0.983 & 38.80 & 0.967 & 0.805       & 2.595    & 3.391        \\
                \hline
                \gr Gating $\rightarrow$ ReLU          & \rt{36.96} & \rt{0.992} & \rt{34.22} & \rt{0.982} & \gt{31.77} &    {0.983} & \rt{38.57} & \rt{0.966} &    {0.805} & \gt{2.585} & \gt{3.384} \\
                \gr \hspace{2.75em} $\rightarrow$ GELU & \rt{37.07} & \rt{0.992} & \rt{34.17} & \rt{0.982} &    {31.60} &    {0.983} & \rt{38.51} & \rt{0.965} &    {0.805} & \gt{2.585} & \rt{3.418} \\
                Sigmoid $\rightarrow$ Hardsig.         &    {38.05} &    {0.993} &    {34.47} &    {0.983} &    {31.53} &    {0.984} &    {38.80} &    {0.967} &    {0.805} &    {2.595} & \gt{3.302} \\
                \hspace{3.35em} $\rightarrow$ Tanh     &    {37.88} &    {0.993} &    {34.53} &    {0.983} &    {31.58} &    {0.984} &    {38.86} &    {0.967} &    {0.805} &    {2.595} & \gt{3.359} \\
                \gr BN $\rightarrow$ LN                & \rt{36.75} & \rt{0.992} & \gt{34.68} &    {0.983} & \rt{30.40} & \rt{0.980} & \rt{38.79} &    {0.967} & \gt{0.803} & \rt{2.608} & \rt{3.643} \\
                \gr \hspace{1.4em} $\rightarrow$ IN    & \rt{35.22} & \rt{0.989} & \rt{33.86} & \rt{0.982} & \rt{30.06} & \rt{0.979} & \rt{38.17} & \rt{0.965} & \gt{0.803} & \rt{2.596} & \rt{3.552} \\
                $k=5$ $\rightarrow$ $k=3$              & \rt{36.99} & \rt{0.992} & \rt{33.55} & \rt{0.980} & \gt{31.93} &    {0.984} & \rt{38.73} &    {0.967} & \gt{0.783} & \gt{2.419} & \gt{3.223} \\
                \hspace{2.4em} $\rightarrow$ $k=7$     & \rt{37.18} & \rt{0.992} & \gt{34.90} & \gt{0.984} & \rt{31.38} &    {0.983} &    {38.82} &    {0.967} & \rt{0.838} & \rt{2.859} & \rt{3.393} \\
                \gr SK $\rightarrow$ Cat               & \rt{36.50} & \rt{0.992} & \rt{31.39} & \rt{0.976} & \rt{29.47} & \rt{0.978} & \rt{38.51} &    {0.967} & \rt{0.824} & \rt{2.801} & \gt{3.001} \\
                \gr \hspace{1.3em} $\rightarrow$ Sum   & \rt{36.33} & \rt{0.991} & \rt{31.36} & \rt{0.977} & \rt{29.48} & \rt{0.978} & \rt{38.57} &    {0.968} & \gt{0.801} & \gt{2.590} & \gt{2.975} \\
                + SE in gConvB                         & \rt{36.61} & \rt{0.992} &    {34.50} &    {0.983} & \rt{31.23} & \rt{0.982} & \gt{38.96} &    {0.968} & \rt{0.852} & \rt{2.606} & \rt{4.787} \\
                + ECA in gConvB                        & \rt{36.76} & \rt{0.992} &    {34.52} & \gt{0.984} & \rt{31.36} &    {0.983} & \gt{38.96} &    {0.968} & \rt{0.806} & \rt{2.606} & \rt{4.253} \\
                \gr Depth $\times$ 2                   & \gt{39.76} & \gt{0.995} & \gt{35.17} & \gt{0.984} & \gt{32.18} & \gt{0.985} & \gt{39.22} & \gt{0.969} & \rt{1.408} & \rt{4.579} & \rt{5.377} \\
                \gr Width $\times$ $\sqrt{2} $         & \gt{38.80} & \gt{0.994} & \gt{34.90} & \gt{0.984} & \gt{31.78} &    {0.984} & \gt{39.03} &    {0.968} & \rt{1.412} & \rt{4.445} & \rt{3.892} \\
                7 Stages $\rightarrow$ 5 Stages        & \rt{36.09} & \rt{0.989} & \rt{32.77} & \rt{0.979} & \rt{30.79} & \rt{0.979} & \rt{37.95} & \rt{0.965} & \gt{0.207} & \gt{1.977} & \gt{2.667} \\
                \hspace{3.35em} $\rightarrow$ 9 Stages &    {37.87} &    {0.993} & \rt{NaN}   & \rt{NaN}   & \gt{31.92} & \gt{0.985} & \rt{NaN}   & \rt{NaN}   & \rt{3.150} & \rt{3.203} & \rt{3.984} \\
                \gr w/o Warmup                         &    {37.70} &    {0.993} &    {34.47} &    {0.983} & \gt{31.75} &    {0.984} & \rt{38.74} &    {0.967} & 0.805 & 2.595 & 3.391 \\
                \gr w/o FrozenBN                       &    {37.83} &    {0.993} &    {34.50} &    {0.983} &    {31.54} &    {0.984} & \rt{38.64} &    {0.967} & 0.805 & 2.595 & 3.391 \\
                \gr w/o Mixed Precision                &    {37.88} &    {0.993} &    {34.49} &    {0.983} &    {31.56} &    {0.983} &    {38.83} &    {0.968} & 0.805 & 2.595 & 3.391 \\
                \hline
            \end{tabular}
            \begin{tablenotes}
                \small
                \item[1] NaN means that the model collapses when training, and although there are valid checkpoints saved, the results are meaningless.
                \item[2] The cost of the nonlinear activation function is ignored and one square root operation is treated as two multiplication operations.
                \item[3] When the Gating is replaced with ReLU or GELU, we keep the two parallel paths but sum them up.
              \end{tablenotes}
            \end{threeparttable}
        }
    \end{center}
    \vspace{-0.75em}
\end{table*}

In our view, ablation studies can better reveal key designs in the model, so we performed ablation studies on all datasets involved in this paper, considering each dataset's characteristics vary so that the ablation studies' results may not be consistent.
We also found that the performance gains of some previous works stem more from the training's randomness.
Therefore, we trained the baseline multiple times and recorded the best and worst performances. 
For models in ablation studies, if their performances fall in between, the corresponding designs are considered not key designs.

\noindent\textbf{Gating Mechanism.}
We validate the improvement by replacing the conventional nonlinear activation function with the gating mechanism.
It can be seen that the introduction of the gating mechanism brings a consistent performance improvement, except on the Haze-4K dataset.
Besides, we found that the performance of ReLU~\cite{nair2010rectified} and GELU~\cite{hendrycks2016gaussian} is comparable, which is different from DehazeFormer's proposal, probably because of the difference in network architecture.
Moreover, the choice of the gating function does not seem to be important, and it is enough to ensure that its output is upper and lower bounded.
We also tried to remove the gating function, but the gradient of the network always explodes during training. 

\noindent\textbf{Normalization Layer.}
We replace the BatchNorm with the LayerNorm and the InstanceNorm~\cite{ulyanov2016instance}.
It can be seen that LayerNorm and InstanceNorm are considerably slower than BatchNorm because they still compute the mean and variance of the feature maps during inference.
And in terms of performance, BatchNorm is slightly better than LayerNorm (especially on RESIDE-IN and Haze-4K datasets) and outperforms InstanceNorm significantly.
Considering the effectiveness and efficiency, BatchNorm should be a better choice than LayerNorm and InstanceNorm for gUNet.
We also tried to remove the normalization layers, but it suffers from gradient explosion.

\begin{figure*}[t]
    \centering
    \includegraphics[width=1.0\textwidth]{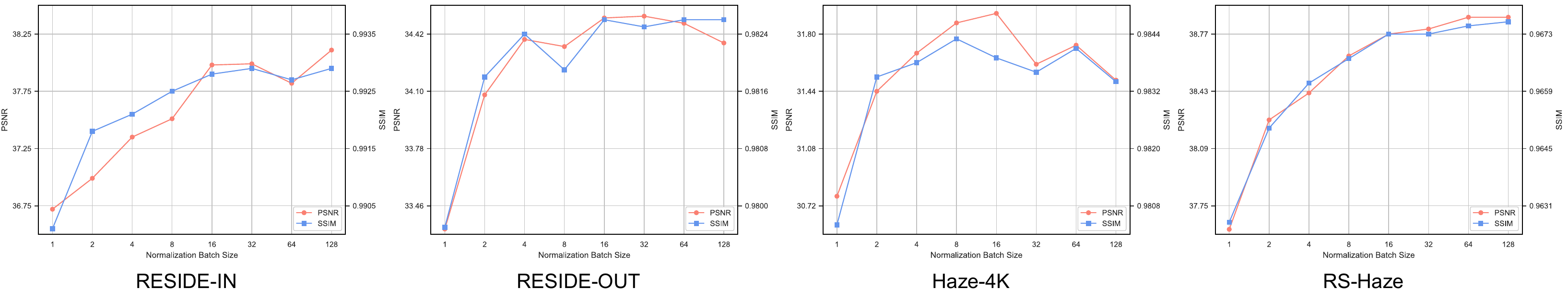}
    \caption{
        Ablation study of training the gUNet-T on different datasets for different normalization batch sizes.
    }
    \label{fig:nbs}
\end{figure*}

\noindent\textbf{Normalization Batch Size.}
If SyncBN is introduced, then the normalization batch size can be equal to the mini-batch size.
Combined GhostBN~\cite{hoffer2017train} with SyncBN, we can set most of the normalization batch sizes needed for this ablation study.
Figure~\ref{fig:nbs} shows the results.
Because the inconsistency of train-test is too large, a small normalization batch size (\emph{e.g.}, 1, 2, or 4) performs poorly.
When the normalization batch size is larger than 8, the model's performance does not change significantly.
Surprisingly, the larger normalization batch size on Haze-4K leads to a performance drop, probably because the Haze-4K dataset is considerably smaller than several others. 
Hence, the small normalization batch size provides more noise to avoid overfitting.

\noindent\textbf{Kernel Size of DWConv.}
ConvNext~\cite{liu2022convnet} proposes that DWConv with too large or too small kernel size will result in performance degradation.
For gUNet, $k=5$ is a good choice.
In addition, we find that larger datasets prefer larger kernel sizes.
Specifically, the sample number of RESIDE-OUT is much larger than other datasets, so the large kernel does not lead to overfitting but improves the expressive capability of the network.
In contrast, Haze-4K has much fewer samples, so setting $k=3$ provides an additional regularization to suppress overfitting.

\noindent\textbf{Fusion Layer.}
If SK Fusion is replaced with the more common summation or concatenation fusion, the gUNet's performance will drop off a cliff.
We believe this is mainly because: 1) SK Fusion dynamically fuses information from different stages, relaxing the constraint of information correlation between stages; 2) SK Fusion can extract global information, which is crucial for estimating global atmospheric light.
Although the overheads of SK Fusion layers in gUNet are negligible, it slightly slows down the network, due to too many element-wise operations~\cite{ma2018shufflenet}.

\noindent\textbf{Channel Attention Mechanism.}
We try to insert the SE module~\cite{hu2018squeeze} and the ECA module~\cite{wang2020eca} after the $\operatorname{PWConv_3}$ of the gConv block, as many image restoration networks~\cite{zhang2018image,anwar2019real,qin2020ffa} do.
Surprisingly, more channel attention modules do not bring a satisfactory improvement. 
On the contrary, they even lead to a significant performance drop on RESIDE-IN and Haze-4K.
We believe SK Fusion serves the purpose of extracting global information and that adding more channel attention modules will not bring any substantial change.
Instead, the gating mechanism can be considered a special attention mechanism, while using two attention modules in the same residual block increases the instability.

\begin{figure*}[t]
    \centering
    \includegraphics[width=1.0\textwidth]{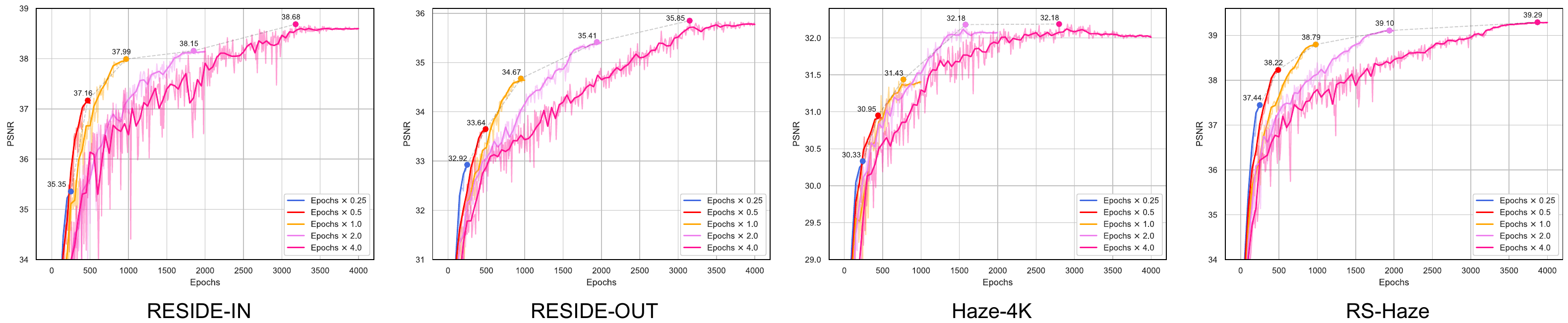}
    \caption{
        Ablation study of training the gUNet-T on different datasets for different epochs.
        The epoch numbers of Warmup and FrozenBN are scaled accordingly.
        The marked PNSR corresponds to the validation performance of the saved best model.
    }
    \label{fig:epochs}
\end{figure*}

\noindent\textbf{Deeper vs. Wider.}
Given the number of parameters and the computational cost, a deeper network performs better than a wider network but at the cost of being slower.
However, a wide network also has a negative effect in that it generates larger intermediate feature maps, consumes more GPU memory, and is detrimental to inference on edge devices.
Therefore, in this paper, we increase the depth of the network to scale the network.

\noindent\textbf{Stage Number.}
We explore the effect of the different number of stages on our model's performance. 
As can be seen, reducing the number of stages lowers the overhead. 
However, it decreases the performance, where the most affected is the number of parameters since the number of parameters is most contributed by the blocks of the intermediate stages.
However, if more stages are introduced, the model only performs comparably to the baseline and generates NaN on two datasets.
We suppose that this is because more stages introduce much more parameters, leading to instability during network training, and the introduction of a more reasonable initialization method helps to overcome this drawback.

\noindent\textbf{Training Stratagies.}
As a result, the performance gains from these technologies are small.
The additional training techniques we use are mainly to improve the model's stability during training and reduce training expenses.
The most important of these techniques for model performance is SyncBN, which is essential for devices where a sufficiently large per-GPU batch size cannot be guaranteed.
Besides, although mixed precision training seems not to impact the model's performance, it makes the training more unstable, as we found that some variants produce NaN in ablation studies.

\noindent\textbf{Training Epochs.}
We found that previous works have not discussed the effect of the number of training epochs on the network performance. 
Figure~\ref{fig:epochs} shows the results of our experiments, with each curve being the PNSR of the model on the validation set during training.
It can be expected that if the network is not overfitted, training longer will bring further improvement to the network.
Besides, we observed overfitting on Haze-4K, which should due to its small sample number.
We note that because training schemes vary across models, researchers tend to replicate the results reported in previous works for comparisons.
However, it is clear that most models are likely not trained to convergence even on the most commonly used RESIDE dataset.
We guess that the performance gains achieved by some of the previous dehazing models are likely to be simply due to that they were trained for a longer time.

\vspace{1em}

\section{Conclusion}

This paper explores the key design of image dehazing networks to achieve performance gains.
Besides the commonly used multi-scale structures and residual learning, efficient utilization of attention mechanisms is the key to improving performance.
Specifically, this paper proposes gUNet, which utilizes the channel attention mechanism in the feature fusion module to extract global information and the gate mechanism to replace the pixel attention and nonlinear activation functions to model spatially varying transmission maps.
We evaluate the performance of gUNet on four image dehazing datasets, and the results show that gUNet is comparable to or even better than state-of-the-art methods with a much smaller overhead.
More importantly, we performed large-scale ablation studies and showed that the performance gains of image dehazing networks go mainly from the attention mechanism, nonlinear activation function, extraction of global information, normalization layer, and the number of training epochs.

\newpage

{\small
\bibliographystyle{ieee_fullname}
\bibliography{egbib}
}

\end{document}